# MOBILITY AND SCIENCE OPERATIONS ON AN ASTEROID USING A HOPPING SMALL SPACECRAFTS ON STILTS

Himangshu Kalita,[*] Stephen Schwartz,[†]
Erik Asphaug[‡] and Jekan Thangavelautham[§]

There are thousands of asteroids in near-Earth space and millions in the Main Belt. They are diverse in physical properties and composition and are time capsules of the early solar system. This makes them strategic locations for planetary science, resource mining, planetary defense/security and as interplanetary depots and communication relays. Landing on a small asteroid and manipulating its surface materials remains a major unsolved challenge fraught with high risk. The asteroid surface may contain everything from hard boulders to soft regolith loosely held by cohesion and very low-gravity. Upcoming missions Hayabusa II and OSIRIS-REx will perform touch and go operations to mitigate the risks of 'landing' on an asteroid. This limits the contact time and requires fuel expenditure for hovering. An important unknown is the problem of getting stuck or making a hard impact with the surface. The Spacecraft Penetrator for Increasing Knowledge of NEOs (SPIKE) mission concept will utilize a small-satellite bus that is propelled using a xenon-fueled ion engine and will contain an extendable, low-mass, high-strength boom with a tip containing force-moment sensors. SPIKE will enable contact with the asteroid surface, where it will perform detailed regolith analysis and seismology as well as penetrometry, while keeping the main spacecraft bus at a safe distance. Using one or more long stilts frees the spacecraft from having to hover above the asteroid and thus substantially reduces or eliminates fuel use when doing science operations. This enables much longer missions that include a series of hops to multiple locations on the small-body surface. We consider a one-legged system, modelled as an inverted pendulum, where the balanced weight is only 10-100 mN. The objective is to balance the spacecraft upon the boom-tip touching the surface. Furthermore, the spacecraft will disengage with the asteroid and hop to another location. The reaction times in the milligravity environment of a km-sized asteroid are much less stringent than the inverted pendulum task on Earth. However, there remain uncertainties with the asteroid surface material, hardness and overall risk posture on the mission. Using this proposed design, we present a preliminary landing system and analyze the implications of GNC on science operations. The proposed spacecraft design and controls approach is a major departure from conventional spacecraft with amphibious capabilities of a lander and flyby vehicle packaged in one.

[*] PhD Student, Space and Terrestrial Robotic Exploration Laboratory, University of Arizona, Tucson, Arizona 85721.
[†] Postdoctoral Research Associate, Lunar and Planetary Laboratory, University of Arizona, Tucson, Arizona 85721.
[‡] Professor, Lunar and Planetary Laboratory, University of Arizona, Tucson, Arizona 85721.
[§] Assistant Professor, Space and Terrestrial Robotic Exploration Laboratory, University of Arizona, Tucson, Arizona 85721.

**INTRODUCTION**

There are one million asteroids in near-Earth space larger than 40 m diameter, and about a thousand bigger than 1 km. These derive from the Main Belt that contains about one million kilometer-sized asteroids and innumerable smaller ones. Asteroids are diverse in physical and dynamical properties and composition. They are time capsules of the early solar system and the planet formation processes. Many are resource-rich containing water, carbon-compounds, iron and platinum group metals. These small bodies are remnants of planet formation, progenitors of meteorites, and are therefore high-value targets identified in the Planetary Science Decadal Survey. Some of these asteroids are potential hazards that may impact Earth.

Asteroid surfaces may contain everything from hard boulders to soft regolith that is loosely held by cohesion and low-gravity. Their global structure may be that of a rubble pile[11] or a fractured monolith, with a tendency of coarser and fewer loose materials as one goes to smaller and smaller size[12]. The unknowns are as basic as if the asteroid behaves as a rigid surface or quicksand[13].

Upcoming missions to asteroids such as Hayabusa II[1] and OSIRIS-REx[2] will perform touch and go operations to mitigate the risks of 'landing' on an asteroid. This limits the contact time with the asteroid and requires significant fuel expenditure for hovering. Landing on a small asteroid and manipulating its surface contents remains a major unsolved challenge fraught with high risk[9,10]. It is a critical engineering challenge that needs to be solved to make surface exploration, resource mining and ambitious plans to setup communication relays and observatories possible.

The extremely low gravity of these smaller asteroids offers some unique opportunities, leading to our proposed approach of a spacecraft on one or more stilts. We presently consider a single stilt, a one-legged lander. Future work shall consider the feasibility of 2 stilts. Stilts may allow hopping on and or off the asteroid using very little energy, eliminating or minimizing the need for combustible chemical propulsion. Instead, conventional electrical thrusters with milli-newtons of thrust might be sufficient to propel on and off small asteroids, especially if a hopping force can be applied by the stilt. This translates into significant cost and risk reduction. This approach opens the opportunity of touching down and visiting multiples sites on an asteroid. Even more ambitious is the possibility of performing subsurface measurements.

We have proposed Spacecraft Penetrator for Increasing Knowledge of NEOs (SPIKE)[3], an ESPA-class solar electric propulsion (SEP) spacecraft. SPIKE is a unique amphibious, lander/flyby spacecraft that would perform multiple landings on an asteroid. It would perform in-situ analysis of the surface and subsurface regolith, in addition to seismic measurements to provide insight on the internal physical structure of the asteroid. The spacecraft touches the asteroid from a probing distance, utilizing an extended boom with science instruments mounted on the tip. Unlike a compact lander [9,10], SPIKE can depart the asteroid after it has completed its science objectives, and with sufficient fuel, can perform tours to multiple asteroid destinations.

In this paper, we present detailed dynamical analysis of the proposed SPIKE spacecraft concept. In particular, we have analyzed the mobility capabilities (descent, landing and hopping) of the spacecraft under expected surface conditions. Our control system accounts for asteroid motion, solar wind and gravity. The spacecraft can be effectively controlled using standard reaction wheels and gimbaled solar-electric Hall thrusters, to enable repeated landings of arbitrary duration. In the following sections, we present related work, the mission concept and preliminary spacecraft design, models of the asteroid environment, spacecraft dynamics and control techniques for descent, landing and hopping followed by conclusion and future work.

**RELATED WORK**

The Apollo astronauts learned that an effective form of mobility on the low-gravity environment of the Moon was through hopping. Our recent studies in low-gravity mobility show hopping to be energy efficient[9,10]. A typical approach to hopping is to use a hopping spring mechanism to overcome large obstacles[17]. One is the Micro-hopper for Mars exploration developed by the Canadian Space Agency[18]. The Micro-hopper has a regular tetrahedron geometry that enables it to land in any orientation at the end of a jump. The hopping mechanism is based on a novel cylindrical scissor mechanism enabled by a Shape Memory Alloy (SMA) actuator. However, the design allows only one jump per day on Mars.

Another technique for hopping developed by Plante and Dubowsky at MIT utilize Polymer Actuator Membranes (PAM) to load a spring. The system is only 18 grams and can enable hopping of Microbots with a mass of 100 g up to a 1 m[19, 20]. Microbots are cm-scale spherical mobile robots equipped with power and communication systems, a mobility system that enables it to hop, roll and bounce and an array of miniaturized sensors such as imagers, spectrometers, and chemical analysis sensors developed at MIT[19, 20]. They are intended to explore caves, lava-tubes, canyons and cliffs.

In our earlier work, we have proposed SphereX[21,22,23,24]. SphereX is a direct descendant of the Microbot platform. SphereX has the same goals as the Microbots, but with the goal of launching fewer robots, that are better equipped with science-grade instruments. SphereX designs have incorporated chemical propulsion[21,22,23] and mechanical torsion springs for hopping[24]. Other techniques for hopping mimic the grasshopper and use planetary gears within the hopping mechanism[25].

NASA JPL and Stanford have developed a planetary mobility platform called "spacecraft/rover hybrid" dubbed "Hedgehog" that relies on internal actuation. With the help of three mutually orthogonal flywheels and external spikes, the platform can perform both long excursions by hopping and short, precise traverses through controlled tumbles[26,27]. Another autonomous microscale surface lander developed is PANIC (Pico Autonomous Near-Earth Asteroid In-Situ Characterizer). PANIC has a shape of a regular tetrahedron with an edge length of 35 cm, mass of 12 kg and utilizes hopping as a locomotive mechanism in microgravity[28].

Mechanical hopping systems can use onboard electrical power, using batteries or PEM fuel cells. PEM fuel cells are especially compelling as techniques have been developed to achieve high specific energy, solid-state fuel storage systems that promise 2,000 Wh/kg[29,30]. Feasibility studies show that the robots can travel 10s of kilometers without recharging. However, for short, focused, simple missions, both the mechanical actuators and systems to power them all impose increased complexity and cost.

It should be noted that all these proposed hopping missions incorporate a group of compact landers. Each lander is limited in its capacity due to its small size and power output, but this compensated through a network working together[9,10,26,27,31,32]. In comparison, SPIKE offers an alternative, of using a single, larger, more-capable lander to perform hopping.

**MISSION CONCEPT**

SPIKE is an amphibious lander/flyby spacecraft propelled using xenon fueled solar-electric Hall thrusters. The onboard instruments would be used to analyze subsurface volatiles and organics and for conducting seismology on asteroids. A potential spacecraft design is based on the JPL Micro Surveyor that has a mass of 75 kg (wet). The science payload will include seismometers, cameras and other instruments that will be designed to access >10 cm beneath the surface of the asteroid. Power is provided by two sets of independently gimballed triple junction solar panels that generate 750 W and a Lithium Ion main battery of 2 kWh. The spacecraft utilizes electric propulsion using Hall Thrusters (JPL MasMi) fueled using Xenon. The spacecraft is estimated to have a delta-V of 5 km/s. The thrusters will be gimballed to desaturate the onboard reaction wheels and to perform landing and take-off maneuvers from the asteroid. The spacecraft attitude determination and control system will utilize Blue Canyon's XACT 50 that includes an integrated start-tracker, IMU and 3 reaction wheels. For communications, it will include JPL's DSN compatible IRIS X-band Radio V2.1 that will permit up to 256 KBps with a rigid 0.3 m parabolic dish. Moreover, the spacecraft has a 3 m rigid deployable truss which holds the science instrument module as shown in Figure 1[3].

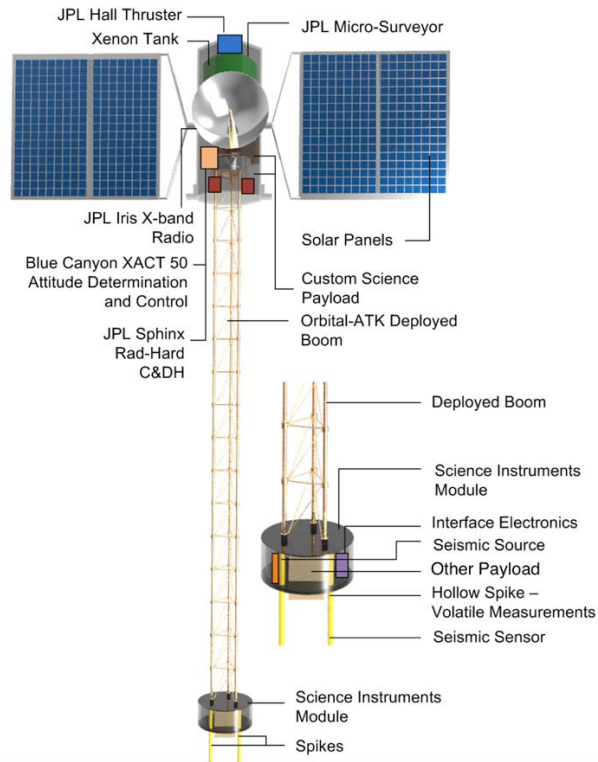

**Figure 1. SPIKE concept fully extended for landing.**

The science mission begins with a survey phase using a camera to determine asteroid rotation state and gravity, and other dynamical characteristics, and to search for the suitable landing sites: regions with regolith, preferably that have been in seasonal shadow for some time as shown in Figure 2 (Concept of Operations). The spacecraft then descends in a precisely guided free-fall, impacting slowly while deploying the extendable boom. The bus comes to rest vertically on top of the science instrument module, deployed at the end of the truss. The base has two penetrators that are pushed into the subsurface to perform science operations. When science operations are complete, the spacecraft disengages by vibrating the regolith to break any cohesion and hops with an initial velocity > 1cm/s. The spacecraft then gets ready to land at a new regolith patch, in another guided free-fall. This is done repeatedly and after the first asteroid investigation is complete, SPIKE travels to a second NEO, and possibly a third[3].

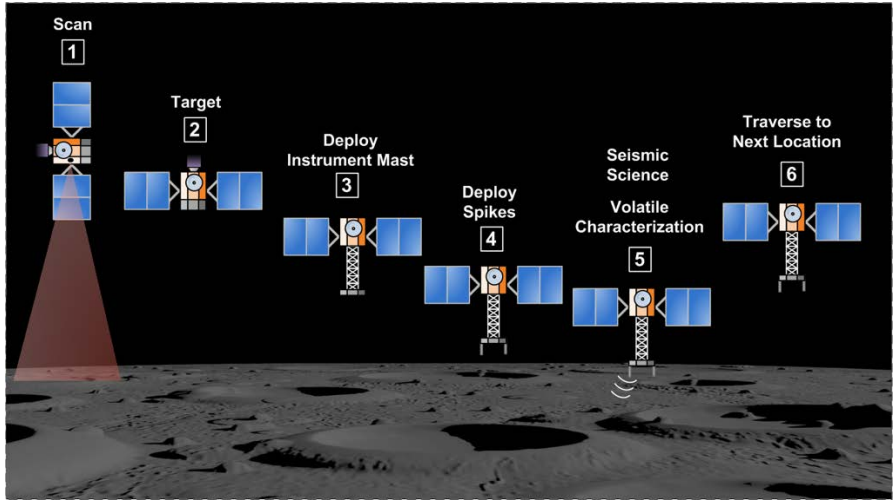

**Figure 2. SPIKE Concept of Operations.**

For the spacecraft to perform detailed regolith analysis and seismology, and penetrometry, it has to make contact with the asteroid's surface. For doing so, the spacecraft is made to follow a general trajectory shape by flying it to one intermediate waypoint. This results into two trajectory segments: the descent phase in which the spacecraft flies to a point directly above the landing site, and the landing phase in which the spacecraft descends under the action of gravity while maintaining its attitude so that it lands vertically on its extended boom. In the next section, we modelled the dynamic environment of the asteroid. The asteroid is modeled as a rigid body that orbits around the sun and its gravitational model and disturbance force model are described. The following section describes the spacecraft dynamic model in the landing site coordinate system. The spacecraft is modeled as a rigid body that approaches the asteroid following the desired trajectory and the simulation results for the descent and landing phase is presented. The hopping dynamics of the spacecraft on its extended boom is then presented with simulation results. Finally, a conclusion based on our research work is presented.

## ASTEROID DYNAMICAL ENVIRONMENT MODELLING

Several works have been done in deriving analytical control laws for descent, landing and surface exploration of spherical and ellipsoidal asteroids. However, as evidenced by remote observations, most asteroids are irregular in shape and are pockmarked by impact craters. An asteroid's shape can be described using a polyhedron, whose surface consists of a series of triangles.

### Gravitational Model

The exterior gravitational potential of a constant density polyhedron was derived analytically by Werner and Scheeres[4] as shown in Eq. (1):

$$V(r) = -\frac{1}{2} G\rho \sum_{e \,\epsilon\, edge} r_e^T E_e r_e \cdot L_e + \frac{1}{2} G\rho \sum_{f \,\epsilon\, face} r_f^T F_f r_f \cdot \omega_f \qquad (1)$$

Where $r_e$ is a vector from the field point to an arbitrary point on each edge, $E_e$ is a dyad defined in terms of the face and edge normal vectors associated with each edge, $L_e$ is a logarithmic term expressing the potential of a 1D straight wire, $r_f$ is a vector from the field point to an arbitrary point on each face, $F_f$ is the outer product of face normal vectors, $\omega_f$ is the solid angle subtended by a face when viewed from the field point.

**Disturbance Forces**

The disturbance forces considered for our analysis are solar radiation pressure and third-body gravitational perturbation modeled as Eq. (2):

$$D = \frac{\eta d \cdot R}{|d|^3} - \frac{\mu}{|R-d|^3}(R-d) \qquad (2)$$

Where η denotes the solar radiation pressure coefficient, $d$ denotes the position vector of the sun from the asteroid, $R$ denotes the position vector of the spacecraft from the asteroid's body fixed coordinate system, and $\mu$ is the product of the gravitational constant and the mass of the sun.

## SPACECRAFT DYNAMICS MODEL

Figure 3 shows the coordinate systems used throughout the paper. $O_a\text{-}X_aY_aZ_a$ is the asteroid's body fixed coordinate system with the origin coinciding with the mass center of the asteroid and $Z_a$ axis coinciding with the asteroid's spin axis. $O_l\text{-}X_lY_lZ_l$ is the landing site coordinate system with the origin coinciding with the preselected landing site and $Z_l$ axis coincides with the vector from the asteroid's center of mass to the landing site. $R_{bs}$ is the position vector from the origin of body fixed coordinate system to the spacecraft, $R_{ls}$ is the position vector from the origin of landing site coordinate system to the spacecraft and $R_{bl}$ is the position vector from the origin of body fixed coordinate system to the origin of landing site coordinate system.

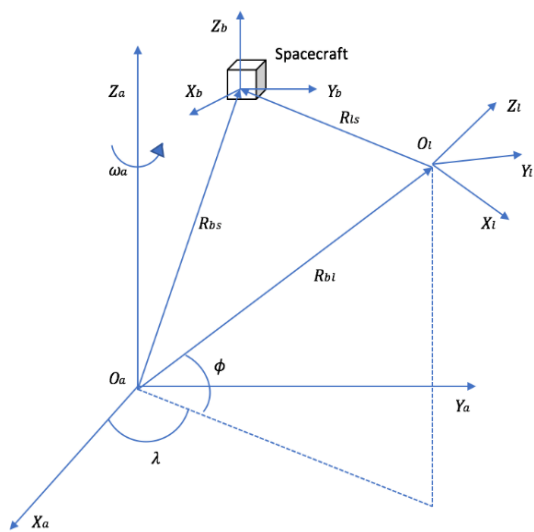

**Figure 3. Coordinate System Definition.**

**Descent and Landing Dynamics and Control**

The dynamic equation of motion of the spacecraft in the asteroid's body fixed coordinate system is shown in Eq. (3):

$$\ddot{R}_{bs} + 2\omega \times \dot{R}_{bs} + \omega \times (\omega \times R_{bs}) + \dot{\omega} \times R_{bs} = U + G + D \tag{3}$$

The asteroid is considered to have a fixed angular velocity vector of $\omega_a$ along the $Z_a$ axis. So, $\omega = [0\ 0\ \omega_a]^T$ and $\dot{\omega} = [0\ 0\ 0]^T$. $U$ denotes the control acceleration, $G$ denotes the gravitational acceleration and $D$ denotes the disturbance acceleration. The dynamic equation of motion in the landing site coordinate system can be written by performing a coordinate transformation from the landing site coordinate system to the body fixed coordinate system. The relationship between the position vectors $R_{bs}$, $R_{ls}$, and $R_{bl}$ is shown in Eq. (4) and (5).

$$R_{bs} = T_l^b R_{ls} + R_{bl} \tag{4}$$

$$T_l^b = \begin{bmatrix} \cos\lambda\sin\phi & -\sin\lambda & \cos\lambda\cos\phi \\ \sin\lambda\sin\phi & \cos\phi & \sin\lambda\cos\phi \\ -\cos\phi & 0 & \sin\phi \end{bmatrix} \tag{5}$$

Hence the equation of motion of the spacecraft in the landing site coordinate system can be written as Eq. (6):

$$\ddot{R}_{ls} + 2\left(T_l^b\right)^{-1}\omega\times\left(T_l^b \dot{R}_{ls}\right) + \left(T_l^b\right)^{-1}\omega\times\left(\omega\times\left(T_l^b R_{ls} + R_{bl}\right)\right) = u + g + d \tag{6}$$

Where $u = \left(T_l^b\right)^{-1}U$, $g = \left(T_l^b\right)^{-1}G$, and $d = \left(T_l^b\right)^{-1}D$ are the control acceleration, gravitational acceleration and disturbance acceleration in the landing site coordinate system respectively.

In the descent phase, the spacecraft is made to follow a general trajectory by flying to a point directly above the intended landing site in time $\tau$. We find a desired acceleration profile $a_d(t)$ that passes through the initial state and the final state[5]. Therefore, we need to solve a two-point boundary-value problem, where the boundary conditions are the initial position and velocity $r_0$, $v_0$, and the final position, velocity, and acceleration $r_f$, $v_f$, and $a_f$. Here, we have considered a quadratic desired acceleration profile, integrating which gives us our desired velocity and desired position profiles as shown in Eq. (7), (8) and (9):

$$a_d(t) = C_0 + C_1 t + C_2 t^2 \tag{7}$$

$$v_d(t) = C_0 t + \frac{1}{2}C_1 t^2 + \frac{1}{3}C_2 t^3 + v_0 \tag{8}$$

$$r_d(t) = \frac{1}{2}C_0 t^2 + \frac{1}{6}C_1 t^3 + \frac{1}{12}C_2 t^4 + v_0 t + r_0 \tag{9}$$

Where, $C_0$, $C_1$ and $C_2$ are coefficients that can be determined by using the boundary conditions (Appendix A). Now, we define position tracking error $e = R_{ls} - r_d$ and velocity tracking error $e_d = \dot{R}_{ls} - v_d$. For the descent phase, we design a PD control law to track the reference position and velocity profiles as shown in Eq. (10):

$$u = -k_p e - k_d e_d \tag{10}$$

Where, $k_p$ and $k_d$ are the proportional and derivative controller gains. For the simulation of the descent phase we have considered the shape model of asteroid Castalia, an asteroid about $1.8 \times 1.4 \times 0.8$ km across. The polyhedron model[14,15] provides the gravitational field in the vicinity of the asteroid and the disturbance forces are computed with respect to the position of the Sun. The position of the landing site is taken as one of the vertices of the polyhedron model whose coordinates w.r.t. the asteroid body fixed frame is $r_l$ = [726, 0, 286] m. The initial position and velocity of the spacecraft are considered to be $r_0$ = [-500, 1000, 1100] m and $v_0$ = [2.2, -1.2, -0.1] m/s w.r.t. the landing site coordinate. The guidance law flies the spacecraft to a point directly over the intended landing site reducing the horizontal component of velocities. Hence, the desired final position and velocity are implemented as $r_f$ = [0, 0, 100] m and $v_f$ = [0, 0, -0.2] m/s w.r.t the landing site reference frame. Figure 4 shows the position and velocity of the spacecraft w.r.t the landing site coordinate frame during the descent phase. Figure 5 shows the control acceleration during the descent phase.

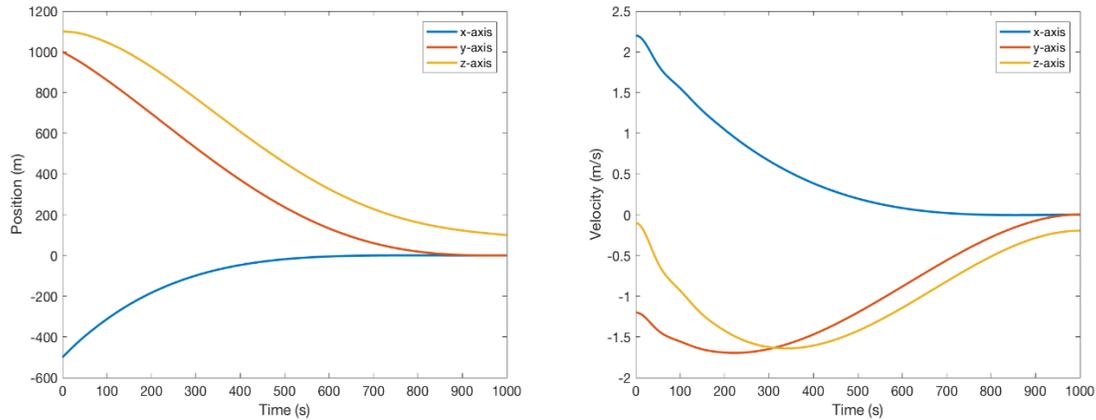

**Figure 4. Position (left) and Velocity (right) of the spacecraft during descent trajectory.**

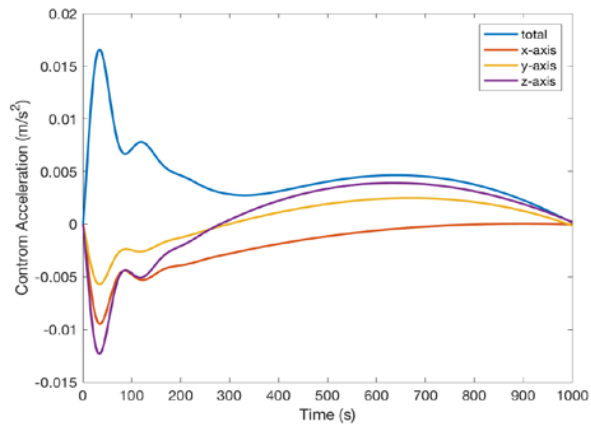

**Figure 5. Control acceleration during the descent phase.**

For the landing phase, the spacecraft descends with an initial vertical velocity under free fall such that the spacecraft hits the landing site with a predefined velocity so that the science instrument module penetrates into the regolith. The final conditions of the descent phase acts as the initial conditions for the landing phase and the spacecraft moves under the action of gravity. However, the spacecraft must maintain the correct attitude so that it lands vertically on its extended boom. The spacecraft with its extended boom is modelled as an inverted pendulum and at any given time, the attitude is determined by applying the rotations from the landing site frame to the body frame of the spacecraft. The orientation of the body-fixed frame with respect to the landing site frame describes the attitude of the spacecraft. The required inputs are computed from the spacecraft's angular-velocity profile. We modelled 3-axis reaction wheels for actuation. If $\omega$ denotes the angular velocity of the body frame with respect to the landing site frame, the attitude dynamics are given by Eq. (11).

$$\dot{\omega} = -J^{-1}(\omega \times J\omega) + \tau_c + \tau_d \tag{11}$$

Where, $J$ is the mass moment of inertia matrix of the spacecraft, $\tau_c$ denotes the control torque, and $\tau_d$ denotes the disturbance torque. The control torque is modelled by a PD control law:

$$\tau_c = -C_p(q-q_d) - C_d(\omega-\omega_d) \tag{12}$$

Where, $C_p$ and $C_d$ are the proportional and derivative controller gains, $q$ and $q_d$ are the actual and desired Euler angles and $\omega$ and $\omega_d$ are the actual and desired angular velocities of the spacecraft. For the simulation of the landing, the initial position and velocity of the spacecraft is taken as the final position and velocity of the descent phase, $r_0 = [0, 0, 100]$ m and $v_0 = [0, 0, -0.2]$ m/s w.r.t. the landing site coordinate system. The initial Euler angles and angular velocities w.r.t the body frame of the spacecraft is considered as $q_0 = [0.1745, -0.3491, 0.3491]$ radians and $\omega_0 = [0.1, 0.2, -0.1]$ rad/s. For calculation of moment of inertia, the spacecraft is considered to have a mass of 50 kg and dimensions 600 × 360 × 360 mm. Each reaction has a mass of 0.75 kg, diameter of 110 mm, height of 38 mm and can apply a maximum torque of 0.025 Nm. Under the action of gravity and the control torques applied by the 3-axis reaction wheels, the spacecraft lands in its final position of $r_f = [3.7, 15.6, 0]$ m with velocity $v_f = [0.02, 0.07, -0.27]$ m/s maintaining its attitude such that $q_f = [0, 0, 0]$ radians and $\omega_f = [0, 0, 0]$ rad/s. Figure 6 shows the position and velocity of the spacecraft during free fall landing. Figure 7 shows the Euler angles and angular velocities of the spacecraft body frame w.r.t the landing site coordinate system during the landing phase. Figure 8 shows the control torque applied by the 3-axis reaction wheel system.

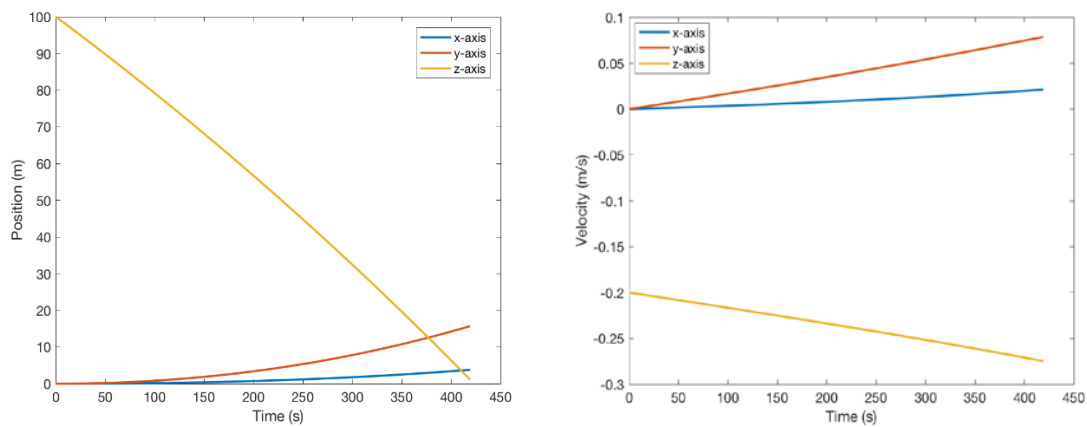

**Figure 6. Position (left) and Velocity (right) of the spacecraft during landing phase.**

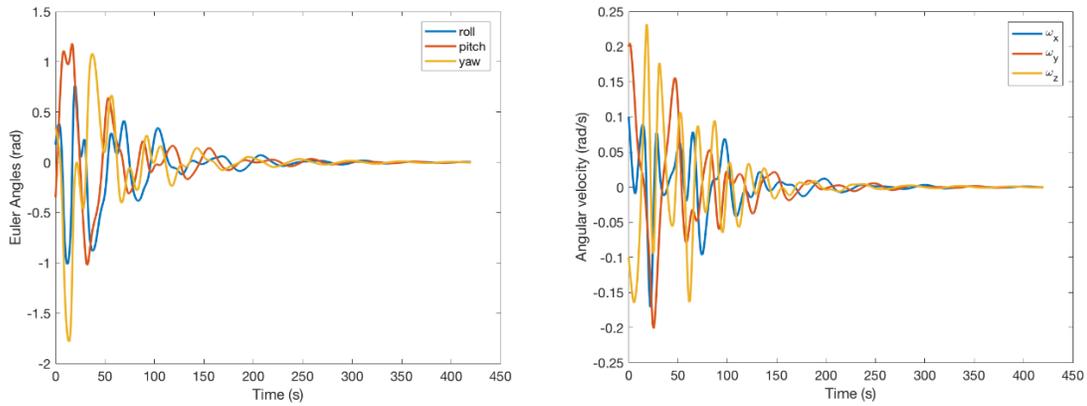

**Figure 7. Euler Angles (left) and Angular Velocity (right) of the spacecraft during landing trajectory.**

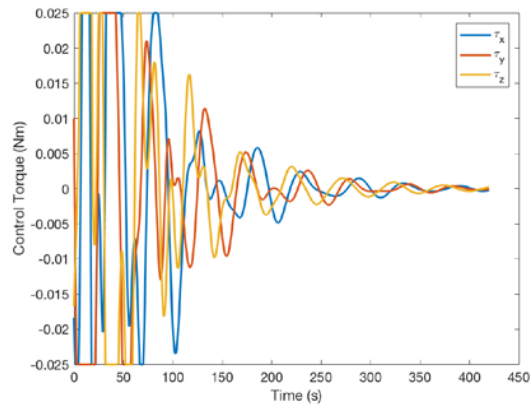

**Figure 8. Control torque applied by the 3-axis reaction wheel system.**

**Hopping Dynamics and Control**

The dynamic equation of motion of the spacecraft in the asteroid's body fixed coordinate system was presented earlier in Eq. 3. The spacecraft hops from rest at position $r_{t0}$ with velocity $v_{t0}$ and impacts at position $r_{tf}$ with velocity $v_{tf}$. The external forces acting on the hopping spacecraft include gravitational forces of the primary body (asteroid) and a tertiary body (Sun) and solar radiation pressure (SRP). The ballistic dynamics of the spacecraft is represented in the rotating body frame, thus introducing effective centrifugal and Coriolis forces. The problem of computing the launch velocity, $v_{t0}$ to intercept a target location, $r_{tf}$ at time $\tau = t_f - t_0$ is the well-known "Lambert orbital boundary-value problem" and solved here.

Efficient numerical solutions have been developed for spherical[6] and perturbed[7] gravity fields. In addition, a "shooting method" has been developed for a polyhedron gravity model[8], which relies on good initial guesses for convergence. Apart from impact targeting, the spacecraft must correct its attitude before landing on its extended boom. The attitude correction is performed with the help of the 3-axis reaction wheels whose dynamics and control is shown in Eq. (11) and (12). In our analysis we have simulated hopping trajectories on asteroid Castalia using initial velocities between 0.1 to 0.45 m/s. It can be observed that a large portion of the trajectories around the asteroid look quite irregular as shown in Figure 9. This observation demonstrates the extreme non-Keplerian behavior of the trajectories around an irregularly shaped minor celestial body[16].

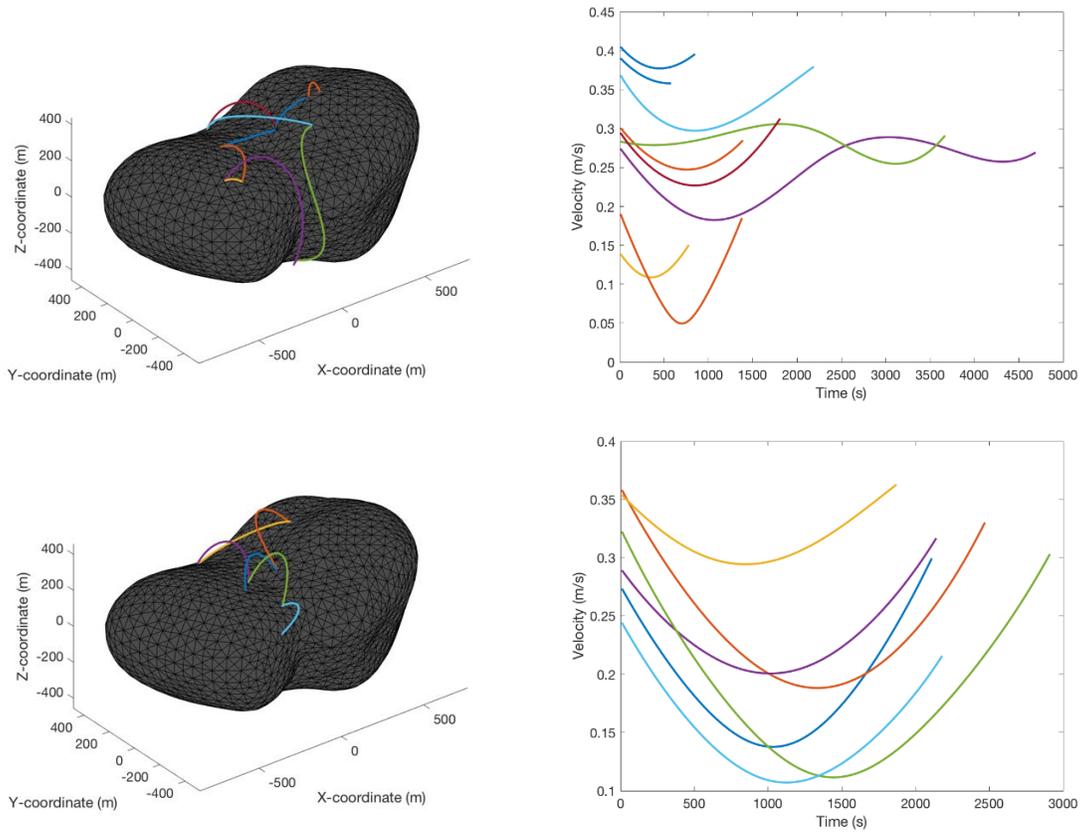

**Figure 9. Multiple hopping trajectories of the spacecraft and their corresponding velocities on the surface of the asteroid.**

## CONCLUSION

In this paper, we presented the GNC capabilities of SPIKE, an amphibious lander-flyby spacecraft for asteroid exploration. We presented the detailed dynamics of the spacecraft w.r.t. to a small asteroid's frame of references. The dynamics and control for a finite-time descent phase is presented with the help of a PD control law that moves the spacecraft directly over the landing site along a desired trajectory. The landing phase is consisting of a free-fall maneuver under the action of gravity. Onboard reaction wheels are used to control the attitude of the spacecraft so that it lands vertically on its extended boom. Using this technique, we presented the principle feasibility of SPIKE performing multiple long-range hops to explore an asteroid. This paper leaves numerous important extensions that are under study and include finding optimal trajectories that minimize fuel consumption and algorithms to maximize area coverage on the asteroid. The path forward shows a compelling, low-cost and innovative method for performing asteroid surface exploration and flybys using small solar electric spacecraft.

# APPENDIX A

Let $\tau$ be the time elapsed for traveling from the initial state to the final target state. We can write Eq. (7), (8) and (9) as:

$$a_\tau = C_0 + C_1\tau + C_2\tau^2 \tag{13}$$

$$v_\tau = C_0\tau + \frac{1}{2}C_1\tau^2 + \frac{1}{3}C_2\tau^3 + v_0 \tag{14}$$

$$r_\tau = \frac{1}{2}C_0\tau^2 + \frac{1}{6}C_1\tau^3 + \frac{1}{12}C_2\tau^4 + v_0\tau + r_0 \tag{15}$$

which can be solved to find the desired coefficients

$$C_0 = a_\tau - 6\left(\frac{v_\tau + v_0}{\tau}\right) + 12\left(\frac{r_\tau - r_0}{\tau^2}\right) \tag{16}$$

$$C_1 = -6\left(\frac{a_\tau}{\tau}\right) + 6\left(\frac{5v_\tau + 3v_0}{\tau^2}\right) - 48\left(\frac{r_\tau - r_0}{\tau^3}\right) \tag{17}$$

$$C_2 = 6\frac{a_\tau}{\tau^2} - 12\left(\frac{2v_\tau + v_0}{\tau^3}\right) + 36\left(\frac{r_\tau - r_0}{\tau^4}\right) \tag{18}$$

To compute the transfer time $\tau$, we can select $\tau$ such that the vertical component of the acceleration profile is a linear function of time which imposes an additional constraint of $C_2 = 0$. Now, the three equations can be solved to find $C_0$, $C_1$ and $\tau$